\documentclass[letterpaper]{article} 
\usepackage{aaai24}  
\usepackage{times}  
\usepackage{helvet}  
\usepackage{courier}  
\usepackage[hyphens]{url}  
\usepackage{graphicx} 
\usepackage{natbib}  
\usepackage{caption} 
\usepackage{colortbl}
\usepackage{makecell}
\usepackage{xcolor}
\usepackage{multirow}
\usepackage{booktabs}

\usepackage{amsmath}
\usepackage{amsfonts}

\usepackage[ruled,linesnumbered]{algorithm2e}

\usepackage{newfloat}
\usepackage{listings}
\usepackage{float}
\DeclareCaptionStyle{ruled}{labelfont=normalfont,labelsep=colon,strut=off} 
\floatstyle{ruled}
\newfloat{listing}{tb}{lst}{}
\floatname{listing}{Listing}
\title{Transforming Graphs for Enhanced Attribute Clustering: An Innovative Graph Transformer-Based Method}
\author{
    \
    Shuo Han\textsuperscript{\rm 1},
    Jiacheng Liu\textsuperscript{\rm 1},
    Jiayun Wu\textsuperscript{\rm 1},
    Yinan Chen\textsuperscript{\rm 1},
    Li Tao\textsuperscript{\rm 1}\thanks{Corresponding author.}
}
\affiliations{
    \textsuperscript{\rm 1}College of Computer and Information Science, Southwest University, Chongqing, China\\

}

\usepackage{bibentry}

\begin{document}

\maketitle

\begin{abstract}
Graph Representation Learning (GRL) is an influential methodology, enabling a more profound understanding of graph-structured data and aiding graph clustering, a critical task across various domains. The recent incursion of attention mechanisms, originally an artifact of Natural Language Processing (NLP), into the realm of graph learning has spearheaded a notable shift in research trends. Consequently, Graph Attention Networks (GATs) and Graph Attention Auto-Encoders have emerged as preferred tools for graph clustering tasks. Yet, these methods primarily employ a local attention mechanism, thereby curbing their capacity to apprehend the intricate global dependencies between nodes within graphs. Addressing these impediments, this study introduces an innovative method known as the Graph Transformer Auto-Encoder for Graph Clustering (GTAGC). By melding the Graph Auto-Encoder with the Graph Transformer, GTAGC is adept at capturing global dependencies between nodes. This integration amplifies the graph representation and surmounts the constraints posed by the local attention mechanism. The architecture of GTAGC encompasses graph embedding, integration of the Graph Transformer within the autoencoder structure, and a clustering component. It strategically alternates between graph embedding and clustering, thereby tailoring the Graph Transformer for clustering tasks, whilst preserving the graph's global structural information. Through extensive experimentation on diverse benchmark datasets, GTAGC has exhibited superior performance against existing state-of-the-art graph clustering methodologies. 
\end{abstract}

\section{Introduction}
Graph representation learning (GRL) is a powerful technique for effectively representing graph-structured data, enabling the extraction of information on graph structure and complex node relationships \cite{SUN}. GRL has found extensive use in various downstream tasks, including node classification \cite{bhagat}, link prediction \cite{liben}, and graph clustering \cite{goyal}.

Graph clustering \cite{tian}is a current area of interest in machine learning, where data is first expressed as a graph and then transformed into a graph partitioning problem \cite{stanton}. Compared to other clustering methods, graph clustering methods have demonstrated superior performance \cite{xie}. These methods can cluster arbitrarily shaped data, overcoming the limitations of traditional clustering methods, which are only effective for clustering convex-shaped data. Graph clustering \cite{tsitsulin} is an essential task in graph analysis \cite{cuzzocrea}, aimed at uncovering the intrinsic structure and interaction patterns of graphs through the clustering analysis of nodes and edges. As an unsupervised learning method, graph clustering algorithms typically rely on a similarity measure to compute the distance or similarity between data points and employ clustering algorithms to group them into distinct clusters. 




In recent years, the attention mechanism, originally stemming from natural language processing (NLP), has found burgeoning applications in graph learning. A notable example is the Graph Attention Network (GAT) \cite{veličković2018graph}, an avant-garde neural network architecture specifically tailored to handle graph-structured data. By employing masked self-attentional layers, GAT transcends the constraints of preceding methods dependent on graph convolutions or their approximations, thereby facilitating the implicit allocation of diverse weights to distinct nodes within a neighborhood. An extension of this concept, GAT-STC \cite{9912369}, integrates Spatial-Temporal Clustering into the Graph Attention Network, thereby augmenting GNN-based traffic predictions in Intelligent Transportation Systems (ITS) through the incorporation of recent-aware and periodic-aware features. Complementing this, Hou et al. \cite{hou2023novel} unveil a clustering algorithm that capitalizes on multi-layer features within graph attention networks, effectively redressing the neglect of shallow features prevalent in deep ensemble clustering.

Graph Attention Auto-Encoders synergize the robust capabilities of attention mechanisms and the principles of unsupervised learning integral to auto-encoders. Their primary function involves distilling a significant representation or encoding from a comprehensive dataset. Uniquely tailored to manage graph-structured data, these auto-encoders extend the application scope of traditional auto-encoders to this specialized domain. Recently, research interest has surged in the area of graph clustering \cite{xu2021graph, wang}, employing Graph Attention Auto-Encoders as a focal tool.

However, both GAT-based and Graph Attention Auto-Encoder-based approaches suffer from a limitation in that they utilize a local attention mechanism \cite{zhao2022eigat}, which may not fully consider the influence of global information on graph clustering tasks. The notion of global information encompasses the comprehensive interconnections and interdependencies that exist among nodes within a graph \cite{li2021global}. Disregarding this crucial information may result in suboptimal clustering outcomes \cite{ostroumova2014global}.



In recent years, the Graph Transformer architecture \cite{yun2019graph} has gained increasing attention in graph representation learning. It naturally overcomes several limitations of graph neural networks (GNNs) \cite{scarselli2008graph} by avoiding their strict structural induction bias and encoding the graph structure  through positional encoding. 
As a generalization of Transformer Neural Network architectures for arbitrary graphs, Graph Transformer extends the key design principles of Transformers \cite{dwivedi2021generalization} from NLP to graphs in general. The Graph Transformer achieves global attention by employing Laplacian eigenvectors to encode node positions and integrating an attention mechanism that enables nodes to attend to every other node in a given graph \cite{rampasek2022GPS}. This powerful feature allows for comprehensive and accurate analysis of the graph structure, leading to superior performance in a variety of graph-based applications.

Similar to the Graph Attention Network (GAT), the Graph Transformer architecture is mainly applicable to graph classification and node-level classification tasks \cite{min2022transformer}. Its attention mechanism is based on computing the similarity between node features and aggregating features from neighboring nodes to update node features \cite{li}. However, this approach is not directly applicable to graph clustering because there is no explicit notion of node similarity or feature aggregation in clustering tasks \cite{jin2021node}.   



To address the challenges associated with attributed graph clustering, we introduce GTAGC (Graph Transformer Auto-Encoder for Graph Clustering). This innovative approach seamlessly integrates the Graph Transformer into a graph autoencoder framework, thereby enhancing its ability to effectively comprehend global relationships and dependencies among graph nodes. The GTAGC model ingeniously amalgamates graph embedding, graph autoencoder, and Graph Transformer architectures, creating a powerful synergy. This fusion enables the capture of both local and global graph information, thereby yielding superior clustering results. The graph embedding component efficiently reduces the dimensionality of the input graph data. Concurrently, the integration of the Graph Transformer within the autoencoder framework bolsters the modeling of long-range node dependencies, underscoring the profound influence of node interconnections. Moreover, the graph autoencoder component preserves the structural subtleties of the graph within the generated embeddings, thereby further enhancing the effectiveness of clustering.

GTAGC crafts an enhanced representation of the original graph, preserving node structural similarities. By reducing differences between the original and projected graphs, it surpasses traditional graph attention network (GAT) limitations, emphasizing global node interconnections for improved clustering accuracy. Our approach blends graph embedding to discern node structural similarities, later applied to clustering \cite{ChenDexiong}. Transitioning between embedding and clustering, GTAGC adeptly leverages the Graph Transformer, maintaining the graph's structural integrity. The main contributions of this paper are listed as follows.
\begin{itemize}
\item We propose a novel graph clustering method called Graph Transformer Auto-Encoder for Graph Clustering \\(GTAGC), which is designed for goal-oriented graph clustering tasks. By ingeniously amalgamating the Graph Auto-Encoder with the Graph Transformer, the GTAGC  successfully harnesses the capability to apprehend global dependencies between nodes.  To the best of our knowledge, this is the first method to effectively utilize the Graph Transformer in graph clustering, providing a unique contribution to the field of graph clustering.


\item Our proposed method combines the Graph Transformer and graph autoencoder techniques to overcome the Graph Transformer's limitation of not being directly applicable to graph clustering. Specifically, we leverage the strengths of graph autoencoder to alternate between graph embedding and clustering, resulting in a more effective approach to graph clustering. 
\item Extensive experimental results on several benchmark datasets demonstrate the superiority of the proposed method against the existing state-of-the-art graph clustering methods.
\end{itemize}

\begin{figure*}[thbp] \centering
    \includegraphics[width=1.08\textwidth,height=1.08\textwidth]{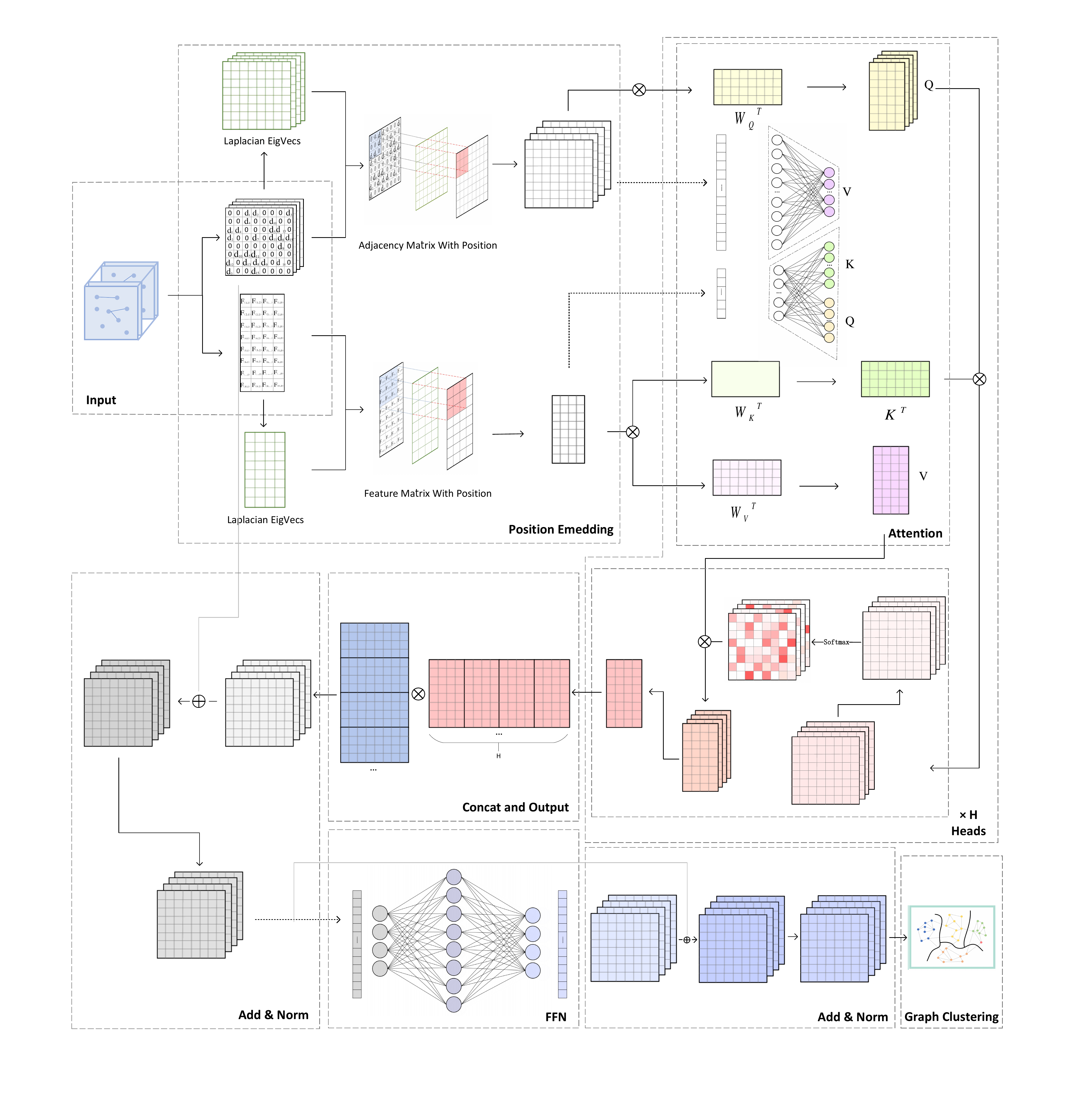}
    \caption{Illustration of the proposed GTAGC. Given a graph G = (V,E), the Laplacian eigenvector matrix is computed based on the graph, which is then utilized for position embedding of the inputs, i.e., the adjacency matrix and the feature matrix. Subsequent to this process, the data is passed through the transformer's H-head self-attention mechanism. The output is obtained by multiplying the result with the matrix O. After a residual connection and normalization, the output undergoes a Feed-Forward Network (FFN) layer, followed by another residual connection and normalization step. After passing through several identical network layers, the final clustering result is acquired.} \label{fig:figure7}
\end{figure*}

\section{Related Work}
This section briefly reviews key advancements in graph and attribute clustering, emphasizing major contributions and their limitations.

\subsection{Graph Embedding}
Graph embedding transforms complex data into vectors or tensors, compacting high-dimensional data into dense vectors \cite{cai2018comprehensive}. Techniques are categorized into matrix decomposition-based and deep learning-based approaches.

In matrix decomposition, Cao et al.'s GraRep model employ a log-transfer probability matrix and Singular Value Decomposition (SVD) \cite{cao2015grarep}, while the HOPE method uses generalized SVD to emphasize asymmetric transitivity in directed networks \cite{ou2016asymmetric}.

Deep learning-based embeddings include SDNE, combining graph similarities \cite{wang2016structural}, DNGR, utilizing random walks and deep auto-encoders \cite{cao2016deep}, VGAE, integrating graph convolutional networks with variational auto-encoders for undirected graph embedding \cite{kipf2016variational}, and Li et al.'s SCDMLGE, a semi-supervised model combining deep metric learning with graph embedding \cite{li2020semi}.

\subsection{Graph Clustering}
Graph clustering groups graph vertices into clusters, focusing on dense connections within clusters and sparse connections between them.

Schaeffer et al. \cite{elisa2007graph} surveyed graph clustering, explaining its definition, methodologies, and evaluation. Zhou et al. \cite{zhou2009je} explored clustering with structural and attribute similarities, introducing the SA Cluster algorithm. Wang et al. \cite{li2023deep} presented a multi-level fusion-based deep graph clustering network, while Guo et al. \cite{guo2022graph} developed an end-to-end framework combining a variational graph autoencoder with generative models.

Recent innovations include contrastive learning in graph clustering. Techniques like Simple Contrastive Graph Clustering (SCGC) \cite{liu2022simple} optimize contrastive learning for deep graph clustering. The Hard Sample Aware Network (HSAN) \cite{liu2022hard} further refines this approach by introducing a holistic similarity measure and a dynamic sample weighting strategy.


\section{Proposed Method}

In this section, we introduce a novel graph clustering method, termed Graph Transformer Auto-Encoder for Graph Clustering (GTAGC). This method integrates the Graph Transformer into a graph autoencoder framework, thereby augmenting its capacity to discern global relationships and dependencies among graph nodes effectively.

\begin{table}[h]
\centering
\caption{Summary of Notations}
\label{tab:notations}
\begin{tabular}{c l}
\hline
\textbf{Notation} & \textbf{Description} \\
\hline

$X$ & Feature matrix of the graph \\
$Z$ & Embedded node feature matrix \\  
$L$ & Number of Graph Transformer layers \\
$\widetilde{X}$ & Smoothed attribute matrix \\
$\widetilde{L}$ & Symmetric normalized graph Laplacian matrix \\
$t$ & Layer number of the Laplacian filter \\
$A$ & Adjacency matrix of the graph \\
$M$ & Binary mask of the graph \\
$D$ & Dimensionality of the input features \\
$F$ & Number of output features \\
$W$ & Weight matrix of the linear transformation \\
$H$ & New feature matrix after linear transformation \\
$a_{global}$ & Global Attention weight matrices \\
$a_{local}$ & Local Attention weight matrices \\
$D$ & Dense attention matrix \\
$A'$ & New adjacency matrix \\
$\alpha_{self}'$ & Normalized attention vector \\
$Q$, $K$, $V$ & Matrices of node feature vectors \\
$d_k$ & Dimensions of the key vectors \\
$Y$ & Output of the clustering layer \\
$\alpha$ & Hyperparameter controlling the balance \\ & between losses \\
\hline
\end{tabular}
\end{table}


\subsection{Main Idea}
The Graph Transformer Auto-Encoder for Graph Clustering (GTAGC) model, proposed in this study, represents a sophisticated deep learning algorithm meticulously designed for efficient graph clustering. Constituted by two principal components, the GTAGC model integrates a Graph Transformer encoder with a dedicated clustering module, synergizing their functionalities to achieve the targeted clustering objectives.


\subsection{Graph Transformer Encoder Module}

The Graph Transformer encoder is designed to take the graph as input and embeds each node into a low-dimensional space while preserving the graph structure. The clustering module then utilizes the embeddings to group the nodes into clusters.

Before encoding, the Laplacian filter \cite{cui2020adaptive} was employed  to perform neighbor information aggregation in the following manner:
\begin{equation}\label{eq:1}
  \widetilde{X} = X(I-\widetilde{L})^{t} \end{equation}

In this equation, $\widetilde{L}$ represents the symmetric normalized graph Laplacian matrix, while $t$ refers to the layer number of the Laplacian filter. Furthermore, $\widetilde{X}$ represents the smoothed attribute matrix.

After that, the Graph Transformer encoder is composed of $L$ Graph Transformer layers, where each layer takes node features $X$ and the adjacency matrix $A$ as input and outputs a new set of node features $Z$. Specifically, the output of the $l\ th$ layer can be mathematically represented as:
\begin{equation}\label{eq:transformer}
\mathbf{Z}^{(l)} = \text{GraphTransformerLayer}(\mathbf{\widetilde{X}}^{(l-1)}, \mathbf{A})^{(1)}
\end{equation}


where the Graph Transformer layer function operates on the input node feature matrix of $layer^{(l-1)}$.


The Graph Transformer encoder  consists of several Graph Transformer (GT) layers with a self-attention mechanism and positional encoding. Each GT layer takes as input a feature matrix $\widetilde{X} \in \mathbb{R}^{N \times D}$, where $N$ is the number of nodes in the graph and $D$ is the dimensionality of the input features. The GT layer also takes in an adjacency matrix $A \in \mathbb{R}^{N \times N}$, which represents the connectivity of the graph, and a binary mask $M \in \mathbb{R}^{N \times N}$, which masks out the self-connections in the graph.

The GT layer first applies a linear transformation to the input features $\widetilde{X}$ using a weight matrix $W \in \mathbb{R}^{D \times F}$, where $F$ is the number of output features. This results in a new feature matrix $H \in \mathbb{R}^{N \times F}$:

\begin{equation}
H = \widetilde{X}W
\end{equation}

The GT layer then computes a global self-attention mechanism on the transformed features $H$. To do this, it learns two attention weight matrices: $a_{self} \in \mathbb{R}^{F \times 1}$ and $a_{local} \in \mathbb{R}^{F \times 1}$, which are used to compute the attention scores for the node itself and its neighboring nodes, respectively. The attention scores for each node are then combined and multiplied with the binary mask $M$ to obtain a dense attention matrix $D \in \mathbb{R}^{N \times N}$:

\begin{equation}
D = Leaky ReLU(\gamma  H a_{local}^T +
H a_{global}^T  )^T \odot M)
\end{equation}

where $\odot$ denotes element-wise multiplication and $Leaky ReLU$ is the LeakyReLU activation function. The coefficient $\gamma$ determines the trade-off between the contribution from neighborhood attention and global attention mechanisms.

The GT layer then applies the attention matrix $D$ to the adjacency matrix $A$ by element-wise multiplication. This results in a new adjacency matrix $A' \in \mathbb{R}^{N \times N}$, where the non-zero entries correspond to the attention scores computed by the self-attention mechanism:

\begin{equation}
A'_{i,j} = \begin{cases}
A_{i,j} D_{i,j} & \text{if } A_{i,j} > 0 \\
-9 \times 10^{15} & \text{otherwise}
\end{cases}
\end{equation}

The GT layer then applies the self-attention mechanism again, this time to the transformed features $H$ to compute an attention vector $a_{self}' \in \mathbb{R}^{N \times 1}$ for each node. The attention scores are then normalized using the softmax function:

\begin{equation}
\alpha_{self}' = Softmax(H a_{self}')
\end{equation}

\begin{algorithm}
\DontPrintSemicolon
\SetAlgoLined
\KwIn{Feature matrix $X \in \mathbb{R}^{N \times D}$, Adjacency matrix $A \in \mathbb{R}^{N \times N}$, Binary mask $M \in \mathbb{R}^{N \times N}$, $\alpha$}
\KwOut{Transformed feature matrix $Z^{(L)}$, Clustering probabilities $\hat{y}$}

Apply Laplacian filter: $\widetilde{X} = X(I-\widetilde{L})^{t}$\;

\For{$l = 1$ \KwTo $L$}{
    Apply Linear Transformation: $H = \widetilde{X}W$\;
    Compute global self-attention mechanism: $D = LeakyReLU(\gamma  H a_{neighs}^T + H a_{global}^T  )^T \odot M$\;
    Update adjacency matrix: $A'_{i,j} = \begin{cases}
    A_{i,j} D_{i,j} & \text{if } A_{i,j} > 0 \\
    -9 \times 10^{15} & \text{otherwise}
    \end{cases}$\;
    Compute self-attention vector: $\alpha_{self}' = Softmax(H a_{self}')$\;
    Compute output features: $H' = (\omega_1 A' + \omega_2 I)\theta H$\;
    Apply FFNN with ReLU activation and batch normalization: $y = BN(\mathrm{W}_{2}FFNN(H')+\mathrm{b}_{2})$\;
    Apply softmax to output: $\hat{y} = \mathrm{Softmax}(y)$\;
    Compute clustering probabilities: $Y = \text{ClusteringLayer}(Z^{(L)})$\;
    Compute Loss: $L = \text{ReconstructionLoss} + \alpha \cdot \text{ClusteringLoss}$\;
}

\Return{$Z^{(L)}$, $\hat{y}$}\;
\caption{Graph Transformer Auto-Encoder for Graph Clustering (GTAGC)}
\end{algorithm}

Finally, the GT layer computes the final output features $H' \in \mathbb{R}^{N \times F}$ by aggregating the transformed features $H$ using the normalized attention matrix $A'$ and the normalized attention vector $a_{self}'$:

\begin{equation}
H' = (\omega_1 A' + \omega_2 I)\theta H
\end{equation}

where $I$ is the identity matrix and $\theta$ is the learnable parameters of the GT layer. The output of the GT layer is then passed through a feedforward neural network (FFNN) with ReLU activation and batch normalization, which is defined as follows:

\begin{equation}
FFNN(x) = \mathrm{ReLU}(BN(\mathrm{W}_{1}x+\mathrm{b}_{1}))
\end{equation}
\begin{equation}
y = BN(\mathrm{W}_{2}FFNN(H')+\mathrm{b}_{2})
\end{equation}

where $W_{1}$, $W_{2}$, $b_{1}$, and $b_{2}$ are the weight matrices and bias vectors of the FFNN, respectively, and BN denotes batch normalization. The final output of the model is obtained by applying a softmax function to y:

\begin{equation}
\hat{y} = \mathrm{Softmax}(y)
\end{equation}


The hyperparameters of the model, including the number of GT layers, the size of the hidden state, the learning rate, and the dropout rate, are selected by grid search on a validation set.


The Graph Transformer layer comprises a self-attentive mechanism and a feedforward neural network. The self-attentive mechanism calculates each node's attention coefficient based on each node's features and its neighbors' features. Then, the feedforward neural network performs a nonlinear transformation of the weighted sum of the neighbors' features to produce the features of the output nodes. The self-attention mechanism can be expressed as follows:

\begin{equation}\label{eq:attention}
\text{Attention}(Q, K, V) = \text{Softmax}\left(\frac{QK^T}{\sqrt{d_k}}\right)V 
\end{equation}

where $Q$, $K$ and $V$ are matrices of node feature vectors and $d_{k}$ are the dimensions of the key vectors. After $L$ Graph Transformer layers, the output node features $Z^{(L)}$ are fed into the clustering module to produce the final clustering results.









\subsection{Clustering Module}

Inspired by DAEGC \cite{wang}, our clustering module operates within an unsupervised learning paradigm, translating node features into a probability set that indicates clustering propensities. Without reliance on ground truth labels, a binary cross-entropy loss function is used to minimize misclassifications, enhancing clustering precision. The output of the clustering layer is formalized as follows:

\begin{equation}
Y = \text{ClusteringLayer}(Z^{(L)})
\end{equation}

where $Y$ signifies the output resulting from the clustering layer function.

The defining equation for the clustering layer function is:

\begin{equation}
\text{ClusteringLayer}(Z) = \frac{\exp(ZW)}{\sum_{j=1}^{k}\exp(ZW_j)}
\end{equation}

In the given equation, $Z$ is a matrix of embedded node features, and $W$ represents a matrix of weights. The clustering layer's output provides clustering probabilities for each node, indicating potential association with different clusters.

The clustering module's loss function is a weighted combination of reconstruction loss, measuring divergence between the input and its reconstruction, and clustering loss, penalizing discrepancies in predicted clustering. The total loss, denoted by $L$, is:

\begin{equation}
L = \text{ReconstructionLoss} + \alpha \cdot \text{ClusteringLoss}
\end{equation}

Here, $\alpha$ is a non-negative hyperparameter controlling the balance between the two losses, optimizing clustering performance during training.

\begin{table*} \centering
\caption{Experimental Results on three Datasets.}
\label{tab:tableres}
\resizebox{\textwidth}{!}{%
\begin{tabular}{l|c|cccccccccccc}
    \toprule
    Dataset & Metric & K-means & Spectral & GraphEncoder & TADW & GAE & VGAE & ARVGE & ARGE & DAEGC & S$^{2}$GC & GC-VAE & GTAGC \\
    \midrule
    \multirow{4}{*}{Citeseer} & ACC & 0.544 & 0.308
 & 0.225 & 0.529 & 0.408 & 0.603	
 & 0.544 & 0.573 & 0.672 & \textcolor{blue}{\textbf{0.691}} &0.666& \textcolor{red}{\textbf{0.708}} \\
    & NMI & 0.312 & 0.090 & 0.033 & 0.320 & 0.174 & 0.343 & 0.261 & 0.350 & 0.397 & \textcolor{blue}{\textbf{0.429}} & 0.409 &\textcolor{red}{\textbf{0.452}} \\
    & F-score & 0.413 & 0.257 & 0.301 & 0.436 & 0.297 & 0.460 & 0.529 & 0.546 & 0.636 & \textcolor{blue}{\textbf{0.647}} & 0.634 & \textcolor{red}{\textbf{0.657}} \\
    & ARI & 0.285 & 0.082 & 0.010 & 0.286 & 0.141 & 0.344 & 0.245 &  0.341 & 0.410 & - & \textcolor{blue}{\textbf{0.415}} & \textcolor{red}{\textbf{0.469}} \\
    \midrule
    \multirow{4}{*}{Cora} & ACC & 0.500 & 0.398 & 0.301 & 0.536 & 0.596 & 0.592 & 0.638 & 0.640 & 0.704 & 0.696 & \textcolor{blue}{\textbf{0.707}}& \textcolor{red}{\textbf{0.717}} \\
    & NMI & 0.317 & 0.297 & 0.059 & 0.366 & 0.397 & 0.408 & 0.450 & 0.449 & 0.528 & \textcolor{blue}{\textbf{0.547}} & 
  0.536 & \textcolor{red}{\textbf{0.540}} \\
    & F-score & 0.376 & 0.332 & 0.230 & 0.401 & 0.415 & 0.456 & 0.627 & 0.619 & 0.682 & 0.658 &  \textcolor{blue}{\textbf{0.695}} & \textcolor{red}{\textbf{0.703}} \\
    & ARI & 0.239 & 0.174 & 0.046 & 0.240 & 0.293 & 0.347 & 0.374 & 0.352 & \textcolor{red}{\textbf{0.496}} & - &  0.482 &  \textcolor{blue}{\textbf{0.489}} \\
    \midrule
    \multirow{4}{*}{Pubmed} & ACC & 0.562 & 0.496 & 0.531 & 0.565 & 0.605 & 0.619 & 0.635 & 0.653 & 0.671 & \textcolor{red}{\textbf{0.710}}&\textcolor{blue}{\textbf{0.682}} & 0.678 \\
        & NMI & 0.262 & 0.147 & 0.210 & 0.224 & 0.232 & 0.216 & 0.232 & 0.248 & 0.263 & \textcolor{red}{\textbf{0.332}}& 0.297 & \textcolor{blue}{\textbf{0.318}} \\
        & F-score & 0.559 & 0.471 & 0.506 & 0.481 & 0.479 & 0.478 & \textcolor{blue}{\textbf{0.678}} & 0.657 & 0.659 & \textcolor{red}{\textbf{0.703}} & 0.669 & 0.664 \\
        & ARI & 0.227 & 0.098 & 0.184 & 0.177 & 0.221 & 0.201 & 0.225 & 0.244 & 0.278 & - & \textcolor{red}{\textbf{0.298}} & \textcolor{blue}{\textbf{0.290}} \\
        \bottomrule
    \end{tabular}
}
\end{table*}

\section{EXPERIMENTS}
In this section, we conducted a series of comprehensive experiments on three widely used datasets \cite{sen2008collective}, namely Citeseer, Cora, and Pubmed, to evaluate the effectiveness of the proposed method.

\subsection{Experimental Settings}
The Graph Transformational Attentional Graph Clustering (GTAGC) model was trained on an NVIDIA 3090 GPU using PyTorch, guided by a framework inspired by the Deep Attentional Embedded Graph Clustering (DAEGC) model \cite{wang}. The training was end-to-end and capped at 200 epochs to align with the DAEGC baseline.

The model's input included a normalized adjacency matrix and a standardized feature matrix for the nodes. The training utilized the Adam Optimizer with a learning rate of 0.001, and early stopping was implemented with patience of 80 epochs to prevent overfitting. If no improvement in validation loss was observed, the best-performing model parameters were restored.


\subsection{Performance Comparison}
To assess the effectiveness and robustness of the proposed Graph Transformer Auto-Encoder for Graph Clustering (GTAGC) was conducted against a range of established methods. These methods include K-means \cite{hartigan1979algorithm}, Spectral Clustering \cite{ng2001spectral}, GraphEncoder \cite{tian2014learning}, TADW \cite{yang2015network}, GAE \cite{tian2014learning}, VGAE \cite{kipf2016variational}, ARVGE \cite{pan2018adversarially}, ARGE \cite{pan2018adversarially}, DAEGC \cite{wang}, S$^{2}$GC \cite{zhu2021simple} and GC-VAE \cite{guo2022graph}. We adopt the code from the original papers for all baseline methods, setting hyperparameters based on author recommendations or applying our own tuning if guidance is lacking. Acknowledging that hardware and operating environment may influence experimental results, we compare our reproduced outcomes with those from the original study, subsequently selecting the optimal values.

The performance of these models was gauged across three distinct datasets: Citeseer, Cora, and Pubmed. Four performance metrics were employed to measure the quality of clustering - accuracy (ACC) \cite{amelio2015normalized}, normalized mutual information (NMI) \cite{amelio2015normalized}, F-score  \cite{li2023deep}, and adjusted rand index (ARI) \cite{robert2021comparing}.

\subsection{Main Results}

We present a comprehensive analysis of the performance of GTAGC on three benchmark datasets, namely Cora, Citeseer, and Pubmed. The results are presented in Table \ref{tab:tableres}, providing detailed insights into the efficacy of GTAGC compared to state-of-the-art graph clustering methods. The values highlighted in red and blue correspond to the highest and second-highest outcomes, respectively.

The presented results shed light on GTAGC's distinguished performance, showcasing its robustness and effectiveness across three widely adopted datasets. Moreover, its commendable results under diverse evaluation metrics, namely ACC, NMI, F-score, and ARI, underline its applicability and prowess in different analytical settings.

\begin{table*}[thb]\centering
    \caption{Ablation study of GTAGC.}
    \label{tab:table4}
    \resizebox{0.72\textwidth}{!}{
    \large
    \begin{tabular}{c|c|cccc|c}
        \toprule
        Dataset & Metric &  \makecell{Baseline} &  \makecell{Laplacian filter} & \makecell{positional encoding} & 
        \makecell{global attention }&
        \makecell{ours}  \\
        
        \midrule
        \makecell{Citeseer} & \makecell{ACC \\ NMI \\ F-score \\ ARI} &  \makecell{0.672 \\ 0.397 \\ 0.636 \\ 0.410} &  \makecell{0.6832\\ 0.4118\\ 0.6366\\ 0.4304}
        
        &  \makecell{0.6688\\ 0.4125\\ 0.6254\\ 0.4173}
        & \makecell{0.6829\\ 0.4253\\ 0.6325\\ 0.4350}
        & \makecell{ \textbf{0.7078}\\\textbf{0.4523}\\\textbf{0.6573}\\\textbf{0.4685}}\\
        \midrule
        \makecell{Cora} & \makecell{ACC \\ NMI \\ F-score \\ ARI} &  \makecell{0.704 \\ 0.528 \\ 0.682 \\ 0.496} &  \makecell{0.6913\\ 0.5154\\ 0.4642\\ 0.6785} &  \makecell {0.6669 \\ 0.5145 \\0.4300 \\ 0.6375} 
        &  \makecell{0.6880 \\ 0.5362 \\0.6594\\0.4574  } & 
        \makecell{\textbf{0.7171} \\
\textbf{0.5402} \\
\textbf{0.7027} \\
\textbf{0.4886} }
        \\
        \bottomrule
\end{tabular}
}
\end{table*}

On the Citeseer dataset, GTAGC excels, standing out as the top performer across every evaluation metric. This demonstrates its ability to adapt to and excel in different performance aspects, providing holistic and effective solutions for graph clustering. These results are particularly notable given the competitiveness of other cutting-edge techniques it is compared with, clearly demonstrating GTAGC's superior potential and efficacy.

Moving to the Cora dataset, GTAGC achieves the highest scores in ACC, NMI, and F-score, affirming its leading position in graph clustering. Although second in ARI, it still delivers a competitive score, underscoring its consistent effectiveness across metrics.

With the Pubmed dataset, GTAGC demonstrates adaptability and resilience. While not leading in ACC, it outperforms many other methods, showcasing its robust performance across various datasets. Its strong ranks in NMI, F-score, and a second-best score in ARI further highlight GTAGC's reliability and consistency in clustering results.



\begin{figure}[thbp]
    \centering
    
    \includegraphics[width=0.232\textwidth]{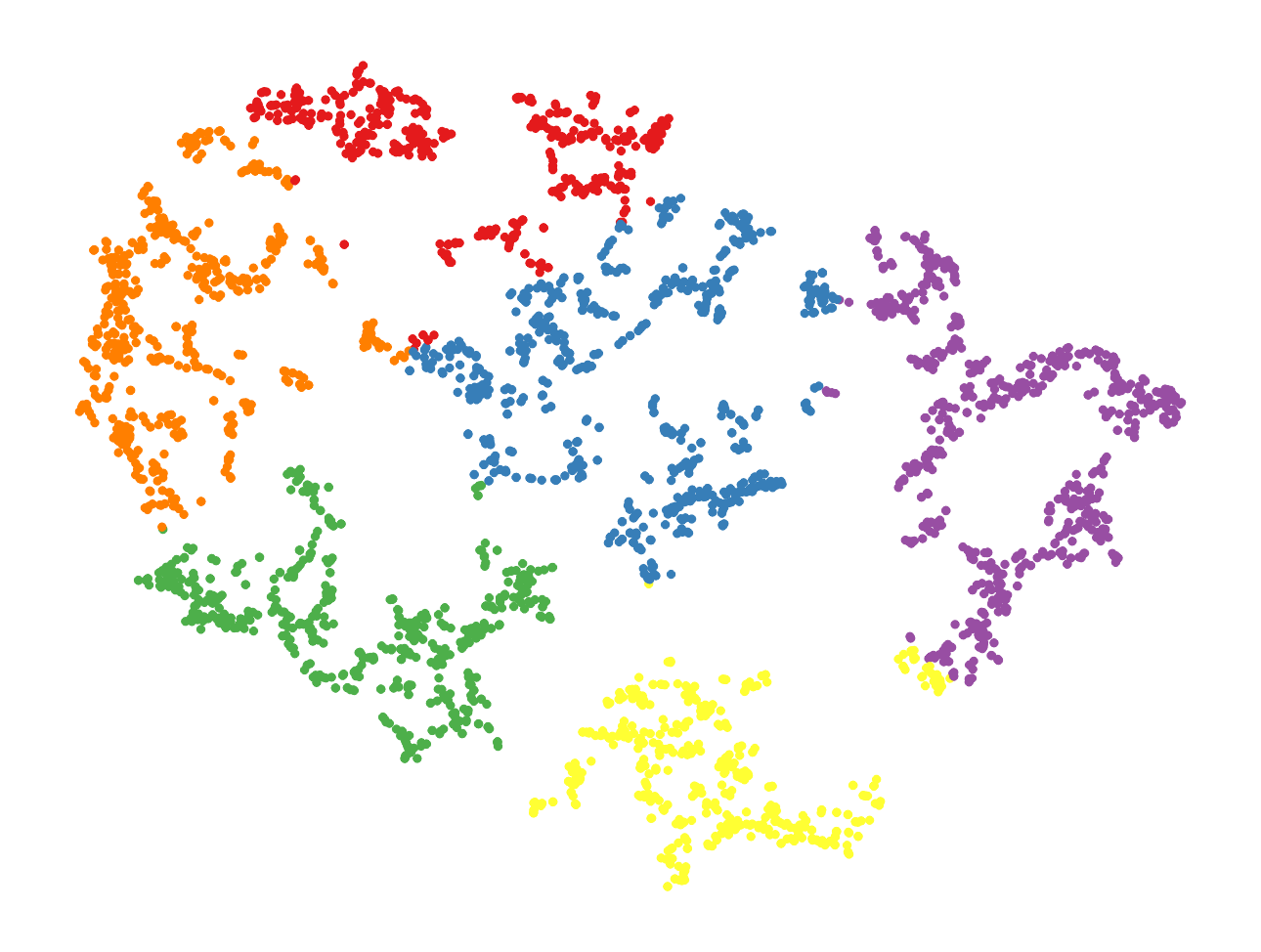}
    \includegraphics[width=0.232\textwidth]{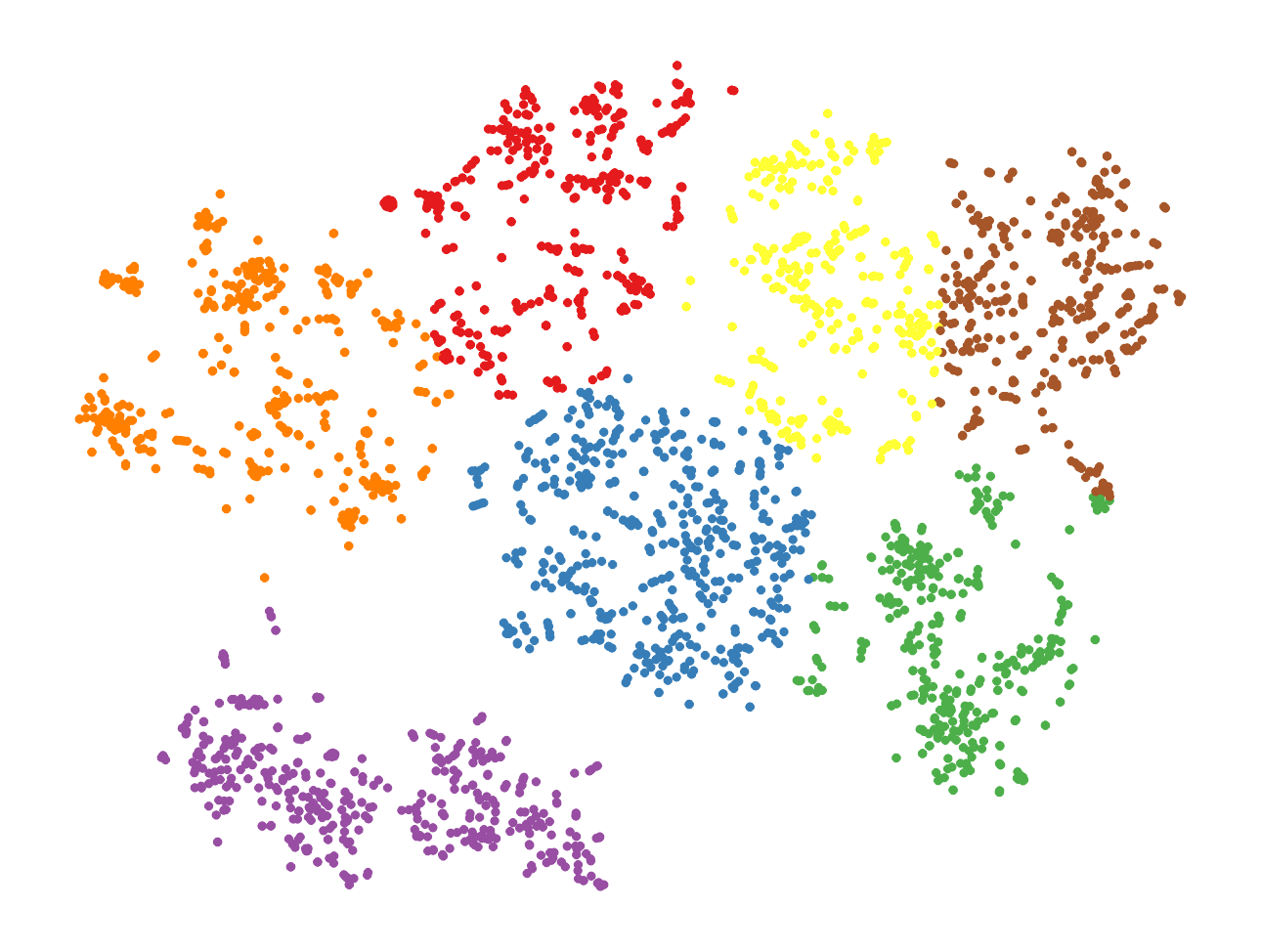}
     \\
    \makebox[0.23\textwidth]{\small (a) Citeseer}
    \makebox[0.23\textwidth]{\small (b) Cora}
    \caption{The 2D visualization of the proposed GTAGC on Citeseer and Cora dataset.}
    \label{fig:fig15}
\end{figure}

In conclusion, these experimental results offer compelling evidence endorsing GTAGC as a trustworthy and robust approach to graph clustering. Its consistently high performance across diverse datasets and metrics, along with its demonstrated ability to compete with or outperform other leading techniques, confidently affirms GTAGC's versatility, reliability, and overall effectiveness.

\subsection{Ablation Studies}
We conducted an ablation study on the Graph Transformational Attentional Graph Clustering (GTAGC) model to assess the individual contributions of each component, using DAEGC as the baseline \cite{wang}. The results are detailed in Table \ref{tab:table4}.

In the Citeseer dataset, the Laplacian filter significantly enhanced accuracy (ACC), normalized mutual information (NMI), and F-score. Incremental improvements were observed with positional encoding, and the global attention mechanism further augmented all metrics, leading the GTAGC model to record the highest scores in ACC, NMI, and F-score.

In the Cora dataset, despite minor declines with the Laplacian filter and positional encoding, the global attention mechanism substantially improved all metrics. Consequently, the GTAGC model excelled in ACC, NMI, and ARI, surpassing existing methods.

In summary, the study emphasizes the vital role of the Laplacian filter, positional encoding, and global attention mechanism in the GTAGC model's superior clustering performance. The combined effect of these components resulted in enhanced accuracy and overall clustering quality, as demonstrated by the GTAGC model's leading performance.


\subsection{Hyper-parameter Analysis}

\begin{figure}[thbp]
    \centering
    
    \includegraphics[width=0.23\textwidth]{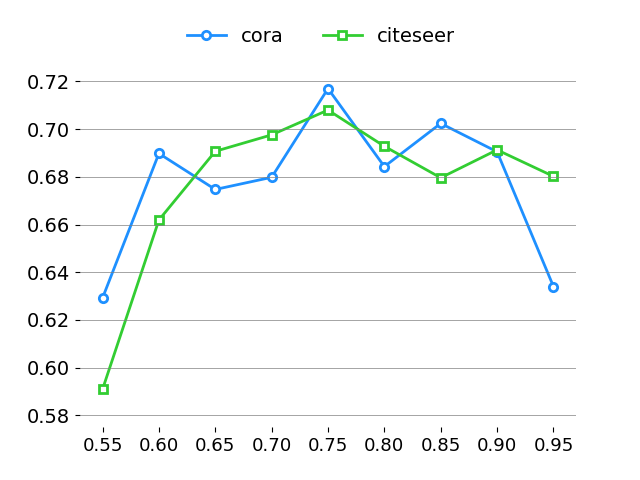}
    \includegraphics[width=0.23\textwidth]{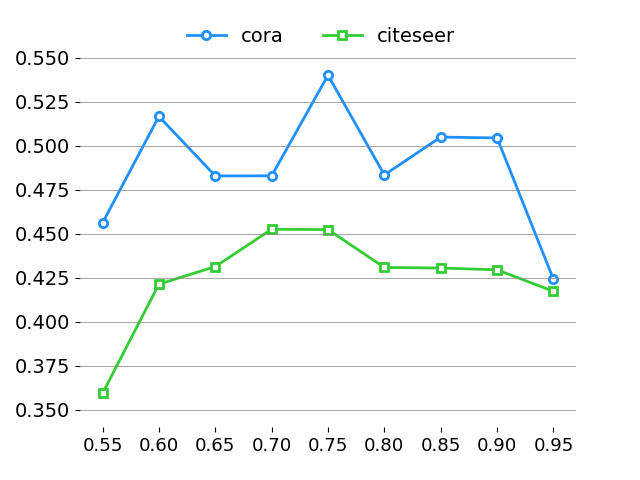}
     \\
    \makebox[0.23\textwidth]{\small (a) ACC}
    \makebox[0.23\textwidth]{\small (b) NMI}
    \caption{Sensitivity analysis of the hyper-parameter $\gamma$ on Citeseer and Cora dataset.}
    \label{fig:fig15}
\end{figure}



We investigated the impact of the hyperparameter $\gamma$, a coefficient governing the equilibrium between local neighborhood and global attention mechanisms. As depicted in Figure \ref{fig:fig15}, clustering outcomes vary with changes in the balance of these attention weights. Notably, optimal clustering is achieved when $\gamma$ is set to 0.75, rendering the local neighborhood attention three times the weight of global attention. This finding underscores the importance of a balanced approach that judiciously integrates both global and local information among nodes, thereby optimizing clustering quality.

\section{Conclusion}

In this paper, we introduce the Graph Transformer Auto-Encoder for Graph Clustering (GTAGC), a pioneering method for attributed graph clustering. Distinct from existing Graph auto-encoder-based approaches, GTAGC enhances flexibility and efficiency in graph clustering through a global attention mechanism. To our knowledge, this constitutes the first integration of Graph Transformer within attributed graph clustering tasks. Future work will explore the incorporation of diverse Graph Transformer variants to further augment the model's capabilities.

\bibliography{aaai24}

\end{document}